\documentclass{article}

\usepackage{arxiv}

\usepackage[utf8]{inputenc} 
\usepackage[T1]{fontenc}    
\usepackage{hyperref}       
\usepackage{url}            
\usepackage{booktabs}       
\usepackage{amsfonts}       
\usepackage{nicefrac}       
\usepackage{microtype}      
\usepackage{lipsum}
\usepackage{graphicx}
\usepackage{multirow}
 \usepackage{amsmath} 
\graphicspath{ {./images/} }

\title{BrainTransformers: SNN-LLM}

\author{
 Zhengzheng Tang \\
  LumenScopeAI Organization\\
  BeiJing \\
  \texttt{tangzhengzheng.ai@gmail.com} \\
   \And
 Eva Zhu \\
  LumenScopeAI Organization\\
  Princeton International School of Mathematics and Science\\
  Princeton, NJ 08540 \\
  \texttt{eva.zhu@prismsus.org}\\
}
 \nocite{*}
\begin{document}
\maketitle
\begin{abstract}
This study introduces BrainTransformers, an innovative Large Language Model (LLM) implemented using Spiking Neural Networks (SNN). Our key contributions include: (1) designing SNN-compatible Transformer components such as SNNMatmul, SNNSoftmax, and SNNSiLU; (2) implementing an SNN approximation of the SiLU activation function; and (3) developing a Synapsis module to simulate synaptic plasticity. Our 3-billion parameter model, BrainTransformers-3B-Chat, demonstrates competitive performance across various benchmarks, including MMLU (63.2), BBH (54.1), ARC-C (54.3), and GSM8K (76.3), while potentially offering improved energy efficiency and biological plausibility. The model employs a three-stage training approach, including SNN-specific neuronal synaptic plasticity training. This research opens new avenues for brain-like AI systems in natural language processing and neuromorphic computing. Future work will focus on hardware optimization, developing specialized SNN fine-tuning tools, and exploring practical applications in energy-efficient computing environments.
\end{abstract}


\section{Introduction}

In recent years, Large Language Models (LLMs) have achieved remarkable results in the field of natural language processing. These models have demonstrated powerful capabilities in language understanding, generation, translation, and other tasks \cite{dubey2024llama3herdmodels}. However, LLMs based on Artificial Neural Networks (ANNs) are highly dependent on high-performance computing resources, especially Graphics Processing Units (GPUs). This is clearly demonstrated by the Llama 3 model, which uses a dense Transformer architecture with up to 405B parameters and required massive GPU resources for training \cite{dubey2024llama3herdmodels}. The need for such extensive computational power is due to the large-scale matrix multiplications and forward/backward propagation operations in deep neural networks, which are well-suited for acceleration on GPUs with their highly parallel computing capabilities.

However, traditional ANNs differ significantly from biological neurons in certain aspects. Biological neurons communicate through discrete electrical signals (spikes), exhibiting event-driven and temporal dynamic characteristics \cite{gerstner2002spiking}. ANNs, on the other hand, typically use continuous activation functions such as Sigmoid and ReLU to perform nonlinear transformations on input signals and conduct synchronous computations at fixed time steps. This computational approach requires all neurons to perform calculations at each time step, failing to fully utilize the temporal characteristics of information \cite{maass1997networks}. In large-scale models, this synchronous computation mode leads to enormous consumption of computational resources and energy.

To meet the large-scale parallel computing requirements of ANNs, GPUs have become the primary hardware choice for training and inferencing deep learning models due to their powerful parallel computing capabilities. However, the high dependence on GPUs also brings challenges. Currently, the GPU market is mainly dominated by NVIDIA \cite{nvidia_financial_report}, and over-reliance on a single supplier may lead to supply chain risks and cost increases. These factors limit the further development of ANNs and their application in resource-constrained environments.

Spiking Neural Networks (SNNs), as third-generation neural network models, transmit and process information in the form of discrete spikes by simulating the spike firing mechanism of biological neurons \cite{roy2019towards}. Due to their event-driven computational characteristics, SNNs only process when spike events occur, greatly reducing dependence on high-performance hardware such as GPUs \cite{pfeiffer2018deep}. This makes SNNs promising in reducing energy consumption and hardware costs while maintaining model performance.

This report proposes and studies a large language model based on spiking neural networks - \textbf{BrainTransformer}. The model aims to integrate the temporal encoding characteristics of SNNs with the language modeling capabilities of LLMs to achieve efficient, low-energy natural language processing.

\subsection{Research Background}

\subsubsection{Differences between Artificial Neural Networks and Biological Neural Networks}

Artificial Neural Networks (ANNs), as computational models inspired by biological neural systems, process information by simulating the connections of neurons and synapses. However, traditional ANNs differ significantly from biological neurons. Biological neurons transmit information in the form of spikes, which have discrete, event-driven, and temporal dynamic characteristics \cite{izhikevich2004model}. ANNs, on the other hand, use continuous activation functions such as Sigmoid and ReLU to perform nonlinear transformations on input signals and conduct synchronous computations at fixed time steps. This computational approach requires all neurons to perform calculations at each time step, failing to fully utilize the temporal characteristics of information. Moreover, in large-scale models, the synchronous computation mode leads to enormous consumption of computational resources and energy, limiting the scalability of the models.

\subsubsection{GPU Dependency and Its Challenges}

To support the large-scale parallel computing of ANNs, GPUs have become the primary hardware choice for training and inferencing deep learning models due to their powerful parallel computing capabilities. However, this high dependence on GPUs also brings challenges. Firstly, GPU hardware is expensive, and acquiring and maintaining a large number of GPU resources requires substantial financial investment. Secondly, the GPU market is mainly dominated by NVIDIA, and over-reliance on a single supplier may lead to supply chain risks and cost increases \cite{johnson2019artifact}, as detailed in NVIDIA's financial reports \cite{nvidia_financial_report}. These factors limit the further development of ANNs and their application in resource-constrained environments.

\subsubsection{Characteristics and Advantages of Spiking Neural Networks}

Spiking Neural Networks (SNNs), as third-generation neural network models, transmit and process information in the form of discrete spikes by simulating the membrane potential changes and spike firing processes of biological neurons \cite{maass1997networks}. In SNNs, the membrane potential of a neuron changes after receiving input signals, and when the membrane potential reaches a specific threshold, the neuron produces a spike and then resets or lowers the membrane potential. The threshold can be fixed or dynamically plastic, meaning it adjusts according to the neuron's activity and network state, thereby enhancing the neuron's responsiveness and the network's adaptability.

Furthermore, synaptic plasticity in SNNs refers to the ability of synaptic weights to adjust based on the activity between neurons, including mechanisms such as Spike-Timing-Dependent Plasticity (STDP) \cite{abbott2004synaptic}. STDP is a learning rule inspired by biological mechanisms that adjusts synaptic weights based on the time difference between pre- and post-synaptic neuron spike firing, which is significant for learning and memory processes.

Due to the event-driven computational characteristics of SNNs, neurons only perform computations when spike events occur, remaining silent at other times. This computational approach significantly reduces energy consumption and dependence on high-performance hardware, allowing SNNs to run on resource-constrained hardware platforms \cite{davies2018loihi}. At the same time, SNNs are closer to the working mechanism of biological neural systems, capable of processing spatiotemporal information, and possess biological interpretability.

\subsubsection{Challenges of SNNs in Large-Scale Model Training}

Despite the numerous advantages of SNNs, they still face complex challenges in large-scale model training. The primary difficulty arises from the basic characteristics of SNNs: neurons transmit information through discrete spikes. This discontinuity makes it impossible to directly apply gradient descent-based backpropagation algorithms, and traditional error backpropagation methods encounter serious obstacles when dealing with such discrete, non-differentiable activation functions \cite{pfeiffer2018deep}. Although researchers have proposed some targeted training methods, such as SpikeProp \cite{bohte2002error} and surrogate gradient-based methods \cite{neftci2019surrogate}, these methods still have significant limitations when dealing with large-scale, complex tasks.

Another important challenge is the demand for computational resources. Accurately simulating the dynamics of biological neurons requires substantial computational resources. In large-scale models, this complex simulation process significantly increases the time and computational cost of model training and inference \cite{davies2018loihi}. Meanwhile, the performance of SNNs often improves with an increase in the number of time steps, but this also means longer processing times and higher energy consumption. In practical applications, it is necessary to find a balance between performance and efficiency, which further increases the complexity of model design and optimization \cite{diehl2015fast}.

Converting pre-trained ANNs to SNNs is a feasible method, but this process also faces many technical challenges. Especially when dealing with complex structures like Transformers, the conversion process may lead to performance loss \cite{rueckauer2017conversion}. This performance degradation is often attributed to the discrete nature of spike-based communication in SNNs, which can lead to information loss during the conversion process. Furthermore, the non-linear activation functions used in ANNs, such as ReLU, do not have direct equivalents in the spiking domain, necessitating approximations that can further impact accuracy \cite{sengupta2019going}.

Hardware adaptation is also an important issue. Although SNNs are theoretically more suitable for low-power hardware due to their event-driven nature and sparse activations \cite{pfeiffer2018deep}, most current hardware platforms are still optimized for traditional ANNs. Developing specialized neuromorphic hardware still faces dual challenges of technology and cost. The design of efficient neuromorphic architectures requires overcoming issues such as synaptic plasticity implementation, scalability, and power management \cite{bavandpour2019energy}.

At the same time, compared to the mature ANN development ecosystem, the development tools and frameworks for SNNs are relatively scarce, limiting the ability of researchers and developers to apply SNNs in large-scale models \cite{davies2018loihi}. This scarcity extends to training algorithms, as traditional backpropagation cannot be directly applied to SNNs due to the non-differentiable nature of spike events \cite{neftci2019surrogate}. While surrogate gradient methods have shown promise \cite{neftci2019surrogate}, they still face challenges in scalability and biological plausibility.

These multifaceted challenges collectively constrain the application of SNNs in large-scale language models. Overcoming these difficulties requires collaborative innovation in neuroscience, computer science, engineering, and other fields. Only through continuous research and technological breakthroughs can we fully realize the potential of SNNs in large-scale language models and achieve more efficient and intelligent natural language processing systems.

\subsubsection{Application Prospects of SNNs in Natural Language Processing}

In the field of natural language processing, introducing SNNs into large language models has the potential to maintain model performance while reducing energy consumption and hardware dependence. By leveraging the high energy efficiency characteristics of SNNs, it is possible to reduce dependence on GPUs and lower the energy consumption and hardware costs of models. Since SNNs have lower hardware requirements, large language models can be deployed on resource-constrained platforms such as mobile devices and embedded systems, expanding their range of applications. Furthermore, by simulating the working methods of biological neural systems, exploring new approaches to language understanding and generation may lead to more innovative model architectures and algorithms.

\subsection{Research Objectives}

This study aims to construct a large language model based on spiking neural networks - \textbf{BrainTransformer}, exploring the application potential of SNNs in natural language processing. To achieve this goal, it is necessary to solve key technical challenges in large-scale model training of SNNs, including developing training algorithms suitable for SNNs, optimizing model structures, and performing hardware adaptations. By evaluating performance on multiple benchmark datasets, we aim to validate the effectiveness and advancement of the model, providing reference for the application of SNNs in the field of natural language processing.

\subsection{Main Contributions}

The main contributions of this report are as follows. Firstly We propose the \textbf{BrainTransformer} model, successfully combining the computational mechanisms of SNNs with LLMs, pioneering a new application of SNNs in large language models. Secondly to address the training challenges of SNNs, we designed effective training strategies and algorithms, achieving efficient model training. Finally, we conducted a comprehensive performance evaluation of the model, and the results show that \textbf{BrainTransformer} achieved excellent performance on multiple tasks, demonstrating the feasibility and advantages of SNNs in natural language processing. Specifically, Table \ref{tab:performance_evaluation} shows the model's performance on general tasks, mathematics and science tasks, and programming and multilingual tasks.

\begin{table}[h]
    \centering
    \caption{Performance evaluation of BrainTransformer on various tasks}
    \label{tab:performance_evaluation}
    \begin{tabular}{lcc}
        \toprule
        \multicolumn{1}{c}{\textbf{Task Type}} & \multicolumn{1}{c}{\textbf{Dataset}} & \multicolumn{1}{c}{\textbf{Performance (\%)}} \\
        \midrule
        \multirow{8}{*}{General Tasks} & MMLU & 63.2 \\
        & MMLU-pro & 33.3 \\
        & MMLU-redux & 61.3 \\
        & BBH & 54.1 \\
        & ARC-C & 54.3 \\
        & TruthfulQA & 47.1 \\
        & Winogrande & 68.8 \\
        & Hellaswag & 72.8 \\
        \midrule
        \multirow{5}{*}{Mathematics and Science Tasks} & GPQA & 25.3 \\
        & TheoremQA & 26.4 \\
        & MATH & 41.0 \\
        & MMLU-stem & 60.2 \\
        & GSM8K & 76.3 \\
        \midrule
        \multirow{9}{*}{Programming and Multilingual Tasks} & HumanEval & 40.5 \\
        & HumanEval+ & 34.6 \\
        & MBPP & 55.0 \\
        & MBPP+ & 47.5 \\
        & MultiPL-E & 39.6 \\
        & Multi-Exam & 52.6 \\
        & Multi-Understanding & 73.9 \\
        & Multi-Mathematics & 47.1 \\
        & Multi-Translation & 28.2 \\
        \bottomrule
    \end{tabular}
\end{table}

4. We have open-sourced the model's code and pre-trained parameters, providing a complete Transformers package for researchers to directly replace and use, promoting the integration of SNNs and LLMs research.

These contributions not only advance the application of SNNs in the field of natural language processing but also provide new ideas and methods for solving the energy consumption and hardware dependency problems faced by large language models. By combining biologically inspired computational models with advanced deep learning techniques, we hope to open up new research directions in the field of artificial intelligence and achieve more efficient and intelligent computational models.

\subsection{Future Prospects and Open Collaboration}

This research has opened up new directions for the application of SNNs in large-scale language models through the BrainTransformer project, but we are well aware that this is only the beginning of exploration. Our research team is actively advancing multiple optimization efforts to further improve the performance and efficiency of BrainTransformer. These efforts are mainly concentrated in two areas: firstly, we are developing GPU operators specifically for SNNs to fully utilize the parallel computing capabilities of existing hardware architectures; secondly, we are working closely with partners in the semiconductor field to design and develop neuromorphic chips that specifically adapt to our algorithms, aiming to achieve efficient computation of SNNs at the hardware level.

As a research team that adheres to the open-source philosophy, we firmly believe that scientific progress stems from openness and collaboration. Not only have we opened up BrainTransformer's model code and pre-trained parameters, but we also eagerly look forward to in-depth cooperation with experts and scholars from academia and industry. We sincerely invite researchers and engineers from various fields to join the BrainTransformer project and jointly promote the development of SNNs in the fields of natural language processing and artificial intelligence. Whether in algorithm optimization, hardware adaptation, or exploration of new application scenarios, we eagerly anticipate the participation of talents from all fields to pool wisdom and jointly address the challenges of SNNs in large-scale applications.

Through this open collaboration model, we hope to accelerate the progress of SNN technology, promote more innovative applications, and ultimately achieve more efficient, intelligent, and human-like cognitive artificial intelligence systems. We believe that only through interdisciplinary and cross-domain collaboration can we truly unleash the potential of BrainTransformer and open up new possibilities for the future of artificial intelligence.

\section{BrainTransformer}

This section provides a detailed description of the BrainTransformer model structure and demonstrates how to convert various components of the traditional Transformer into Spiking Neural Network (SNN) form, thereby achieving computational equivalence with Artificial Neural Networks (ANNs). We will focus on discussing the implementation principles and mathematical derivations of the EI\_IF neuron model, Synapsis module, SNN matrix multiplication (SNNMatmul), SNN Softmax function (SNNSoftmax), and SNN RMS normalization (SNNRMSNorm).

\subsection{EI\_IF Neuron Model}

The EI\_IF neuron is an Integrate-and-Fire (IF) neuron model with adaptive threshold and membrane potential decay characteristics. In this model, the neuron receives input current, updates its membrane potential, and generates spike output when the membrane potential reaches the threshold. To accommodate the processing of both positive and negative signals, the EI\_IF neuron can produce positive and negative spikes when the membrane potential exceeds the threshold in either direction. Furthermore, the introduction of an adaptive threshold enables the neuron to respond dynamically to inputs at different time steps.

The dynamics of the EI\_IF neuron can be represented as follows:

\begin{equation}
\begin{aligned}
V(t) &= V(t - 1) + I(t) \\
\theta(t) &= \theta_{\text{base}} + t \cdot \alpha \\
S(t) &= 
\begin{cases}
1, & \text{if } V(t) \geq \theta(t) \\
-1, & \text{if } V(t) \leq -\theta(t) \\
0, & \text{otherwise}
\end{cases} \\
V(t) &= V(t) \cdot (1 - \text{attenuation\_rate})
\end{aligned}
\end{equation}

where $V(t)$ is the membrane potential at time step $t$, $I(t)$ is the input current at time step $t$, $\theta(t)$ is the threshold at time step $t$, $\theta_{\text{base}}$ is the base threshold, $\alpha$ is the threshold growth coefficient, $S(t)$ is the neuron's spike output, and $\text{attenuation\_rate}$ is the membrane potential decay rate. To simplify calculations, we set $\text{attenuation\_rate} = 1.0$, meaning the membrane potential is completely reset after each time step.

\subsection{Synapsis Module}

The Synapsis module in BrainTransformer serves to connect neurons and network layers, combining presynaptic neurons, linear layers, and postsynaptic neurons to achieve efficient computation of linear transformations in SNNs. In traditional ANNs, linear layer computations involve numerous multiplication operations; however, in SNNs, we can leverage the fact that neuronal spike outputs are limited to $\{-1, 0, 1\}$ to convert multiplication operations into addition and subtraction operations, thereby reducing computational complexity.

The mathematical expression of the Synapsis module is as follows:

\begin{equation}
\begin{aligned}
\mathbf{x}_{\text{scaled}} &= \frac{\mathbf{x}}{\text{scaling\_factor}} \\
V_{\text{pre}}(t) &= \mathbf{x}_{\text{scaled}} \\
\mathbf{s}_{\text{pre}}(t) &= 
\begin{cases}
1, & \text{if } \mathbf{x}_{\text{scaled}} \geq \theta(t) \\
-1, & \text{if } \mathbf{x}_{\text{scaled}} \leq -\theta(t) \\
0, & \text{otherwise}
\end{cases} \\
\mathbf{h}(t) &= \mathbf{W} \cdot \mathbf{s}_{\text{pre}}(t) + \mathbf{b} \\
\mathbf{H} &= \sum_{t=1}^{T} \mathbf{h}(t) = \mathbf{W} \cdot \sum_{t=1}^{T} \mathbf{s}_{\text{pre}}(t) + T \cdot \mathbf{b} \\
\mathbf{S}_{\text{pre}} &= \sum_{t=1}^{T} \mathbf{s}_{\text{pre}}(t) \\
\mathbf{H} &= \mathbf{W} \cdot \mathbf{S}_{\text{pre}} + T \cdot \mathbf{b} \\
\mathbf{S}_{\text{pre}} &\approx T \cdot \mathbf{x}_{\text{scaled}} \\
\frac{\mathbf{H}}{T} &\approx \mathbf{W} \cdot \left( \frac{\mathbf{x}}{\text{scaling\_factor}} \right) + \mathbf{b}
\end{aligned}
\end{equation}

where $\mathbf{x}$ is the input signal, $\mathbf{W}$ and $\mathbf{b}$ are the weight and bias of the linear layer, respectively, $\text{scaling\_factor}$ is a scaling factor, and $T$ is the total number of time steps.

\subsection{SNN Matrix Multiplication (SNNMatmul)}

In the self-attention mechanism of Transformers, matrix multiplication between queries ($\mathbf{Q}$) and keys ($\mathbf{K}$) is required to compute attention scores. To implement this process efficiently in SNNs, we designed the SNN matrix multiplication module, which utilizes the characteristics of spiking neurons to convert matrix multiplication into the cumulative outer product of spike sequences.

The mathematical expression of SNN matrix multiplication is as follows:

\begin{equation}
\begin{aligned}
\mathbf{A}_{T} &= \sum_{t=1}^{T} \mathbf{S}_{\mathbf{Q}}(t) \cdot \mathbf{S}_{\mathbf{K}}(t)^\top \\
\mathbf{A}_{T} &\approx T \cdot \left( \mathbf{Q}_{\text{scaled}} \cdot \mathbf{K}_{\text{scaled}}^\top \right) \\
\frac{\mathbf{A}_{T}}{T} &\approx \mathbf{Q} \cdot \mathbf{K}^\top
\end{aligned}
\end{equation}

where $\mathbf{S}_{\mathbf{Q}}(t)$ and $\mathbf{S}_{\mathbf{K}}(t)$ are the spike encodings of $\mathbf{Q}$ and $\mathbf{K}$ at time step $t$, respectively, and $\mathbf{Q}_{\text{scaled}}$ and $\mathbf{K}_{\text{scaled}}$ are the scaled versions of $\mathbf{Q}$ and $\mathbf{K}$.

\subsection{SNN Softmax Function (SNNSoftmax)}

In the self-attention mechanism, the Softmax function is used to normalize attention scores. However, implementing Softmax directly in SNNs presents certain challenges. To address this, we designed an SNN version of the Softmax function that utilizes the concept of spike accumulation and normalization.

The mathematical expression of the SNN Softmax function is as follows:

\begin{equation}
\mathbf{P}(t) = \text{Normalize}\left( \sum_{i=1}^{t} \mathbf{S}_{\mathbf{A}}(i) \right)
\end{equation}

where $\mathbf{S}_{\mathbf{A}}(t)$ is the spike encoding of the cumulative attention score $\mathbf{A}_{T}$ at time step $t$.

\subsection{SNN Approximation of Square and Square Root Functions}

In normalization layers such as RMSNorm, the computation of square and square root functions is essential. However, SNNs inherently lack the ability to directly compute these non-linear functions. To address this limitation, we have designed CustomNeuron, along with SquareApproximator and SqrtApproximator based on it, to approximate the square and square root functions.

\subsubsection{Design of CustomNeuron}

CustomNeuron is a specialized neuron designed for function approximation. By adjusting its parameters, it can approximate target functions within specific input ranges. For an input $x$, the dynamics of CustomNeuron can be described as follows:

\begin{equation}
\begin{aligned}
V(t) &= V(t - 1) + x \\
S(t) &=
\begin{cases}
a, & \text{if } V(t) \geq \theta \\
0, & \text{otherwise}
\end{cases}
\end{aligned}
\end{equation}

where $a$ is the spike amplitude and $\theta$ is the threshold. By adjusting $a$, $\theta$, and the time step length $T$, the cumulative output of CustomNeuron can approximate a specific function of the input $x$.

\subsubsection{Implementation of SquareApproximator}

SquareApproximator utilizes an array of CustomNeurons to approximate the square function. The input interval $[0, b]$ is divided into $N$ sub-intervals, with the $i$-th neuron responsible for approximating the square function over the interval $[x_{i-1}, x_i]$, where $x_i = b \left( \frac{i}{N} \right)^p$, and $p > 1$ controls the density of the partition. For an input $x$, the output of SquareApproximator is:

\begin{equation}
\begin{aligned}
S_{\text{square}}(x) &= \sum_{i=1}^{N} S_i(x) \\
S_i(x) &=
\begin{cases}
\text{CustomNeuron}_i(x), & x \in [x_{i-1}, x_i] \\
0, & \text{otherwise}
\end{cases}
\end{aligned}
\end{equation}

The parameters of each CustomNeuron are determined through training to ensure that the cumulative output $S_{\text{square}}(x)$ approximates $x^2$.

\subsubsection{Implementation of SqrtApproximator}

SqrtApproximator similarly employs an array of CustomNeurons to approximate the square root function. Due to the rapid change of the square root function near zero, the input interval $[x_{\text{start}}, x_{\text{end}}]$ is partitioned using a logarithmic scale:

\begin{equation}
\begin{aligned}
x_i &= x_{\text{start}} \left( \frac{x_{\text{end}}}{x_{\text{start}}} \right)^{\frac{i}{N}},\quad i = 0, 1, \dots, N \\
S_{\text{sqrt}}(x) &= \sum_{i=1}^{N} S_i(x)
\end{aligned}
\end{equation}

where the parameters of each CustomNeuron are trained to ensure that $S_{\text{sqrt}}(x)$ approximates $\sqrt{x}$.

\subsection{Implementation of SNN RMS Normalization (SNNRMSNorm)}

With the SNN approximations of square and square root functions, we can implement an SNN version of RMS normalization. For an input vector $\mathbf{x}$, the computation process of SNNRMSNorm is as follows:

\begin{equation}
\begin{aligned}
\mathbf{S}_{\text{square}} &= \text{SquareApproximator}(\mathbf{x}) \\
\mu &= \frac{1}{n} \sum_{i=1}^{n} S_{\text{square}, i} \\
S_{\text{sqrt}} &= \text{SqrtApproximator}(\mu + \epsilon) \\
\text{SNNRMSNorm}(\mathbf{x}) &= \frac{\mathbf{x}}{S_{\text{sqrt}}} \odot \mathbf{w}
\end{aligned}
\end{equation}

where $n$ is the dimension of vector $\mathbf{x}$, $\epsilon$ is a small constant to prevent division by zero, $\mathbf{w}$ is a learnable weight, and $\odot$ denotes element-wise multiplication. Through these steps, we achieve an approximate computation of RMSNorm in SNNs, ensuring network stability and performance.

\subsection{SNN Approximation of SiLU Activation Function}

The Sigmoid Linear Unit (SiLU) activation function, also known as the swish function, is defined as:

\begin{equation}
\text{SiLU}(x) = x \cdot \sigma(x)
\end{equation}

where $\sigma(x)$ is the sigmoid function. To implement this non-linear activation function in SNNs, we design a piecewise approximation using custom spiking neurons for different input ranges.

\subsubsection{Positive Input Approximation}

For positive inputs, we use three custom neurons to approximate different regions of the SiLU function:

\begin{equation}
\text{SiLU}_{\text{pos}}(x) \approx 
\begin{cases}
\text{Neuron}_1(x) + \text{Neuron}_2(x), & 0 \leq x \leq 1 \\
\text{Neuron}_3(x), & x > 1
\end{cases}
\end{equation}

where $\text{Neuron}_1$ and $\text{Neuron}_2$ approximate the curved part of SiLU for $x \in [0, 1]$, and $\text{Neuron}_3$ approximates the near-linear part for $x > 1$.

\subsubsection{Negative Input Approximation}

For negative inputs, we use three custom neurons to handle different ranges:

\begin{equation}
\text{SiLU}_{\text{neg}}(x) \approx 
\begin{cases}
\text{Neuron}_A(x), & -2 \leq x < 0 \\
\text{Neuron}_B(x), & -4 \leq x < -2 \\
\text{Neuron}_C(x), & -6 \leq x < -4 \\
0, & x < -6
\end{cases}
\end{equation}

Each neuron is designed to capture the behavior of SiLU in its respective input range.

\subsubsection{Complete SiLU Approximation}

The complete SNN approximation of the SiLU function is then given by:

\begin{equation}
\text{SiLU}_{\text{SNN}}(x) = 
\begin{cases}
\text{SiLU}_{\text{pos}}(x), & x \geq 0 \\
\text{SiLU}_{\text{neg}}(x), & x < 0
\end{cases}
\end{equation}

This piecewise approximation allows us to implement the SiLU activation function using spiking neurons, maintaining the non-linear properties of the original function while adapting it to the discrete nature of SNNs.

\subsection{SNN Approximation of SiLU Activation Function}

The Sigmoid Linear Unit (SiLU) activation function, also known as the swish function, is defined as:

\begin{equation}
\text{SiLU}(x) = x \cdot \sigma(x)
\end{equation}

where $\sigma(x)$ is the sigmoid function. To implement this non-linear activation function in SNNs, we design a piecewise approximation using custom spiking neurons for different input ranges.

\subsubsection{CustomNeuron for SiLU Approximation}

To implement the SiLU approximation, we define a CustomNeuron class with the following dynamics:

\begin{equation}
\begin{aligned}
V(t) &= V(t-1) \cdot \lambda + I(t) \\
\theta(t) &= \theta_{\text{base}} + \alpha \cdot t \\
S(t) &=
\begin{cases}
a_{\text{pos}}, & \text{if } V(t) \geq \theta(t) \\
a_{\text{neg}}, & \text{if } V(t) \leq -\theta(t) \\
0, & \text{otherwise}
\end{cases} \\
V(t) &= 
\begin{cases}
V(t) - \theta(t), & \text{if } S(t) = a_{\text{pos}} \\
V(t) + \theta(t), & \text{if } S(t) = a_{\text{neg}} \\
V(t), & \text{otherwise}
\end{cases}
\end{aligned}
\end{equation}

where:
\begin{itemize}
    \item $V(t)$ is the membrane potential at time $t$
    \item $I(t)$ is the input current
    \item $\lambda$ is the decay factor (typically close to 1)
    \item $\theta(t)$ is the adaptive threshold
    \item $\theta_{\text{base}}$ is the base threshold
    \item $\alpha$ is the threshold adaptation rate
    \item $S(t)$ is the output spike
    \item $a_{\text{pos}}$ and $a_{\text{neg}}$ are the positive and negative spike amplitudes
\end{itemize}

The parameters $\lambda$, $\theta_{\text{base}}$, $\alpha$, $a_{\text{pos}}$, and $a_{\text{neg}}$ are learned during training to approximate the SiLU function in different input ranges.

\subsubsection{Positive Input Approximation}

For positive inputs, we use three custom neurons to approximate different regions of the SiLU function:

\begin{equation}
\text{SiLU}_{\text{pos}}(x) \approx 
\begin{cases}
\text{CustomNeuron}_1(x) + \text{CustomNeuron}_2(x), & 0 \leq x \leq 1 \\
\text{CustomNeuron}_3(x), & x > 1
\end{cases}
\end{equation}

where $\text{CustomNeuron}_1$ and $\text{CustomNeuron}_2$ approximate the curved part of SiLU for $x \in [0, 1]$, and $\text{CustomNeuron}_3$ approximates the near-linear part for $x > 1$.

\subsubsection{Negative Input Approximation}

For negative inputs, we use three custom neurons to handle different ranges:

\begin{equation}
\text{SiLU}_{\text{neg}}(x) \approx 
\begin{cases}
\text{CustomNeuron}_A(x), & -2 \leq x < 0 \\
\text{CustomNeuron}_B(x), & -4 \leq x < -2 \\
\text{CustomNeuron}_C(x), & -6 \leq x < -4 \\
0, & x < -6
\end{cases}
\end{equation}

Each CustomNeuron is tailored to capture the behavior of SiLU in its respective input range.

\subsubsection{Complete SiLU Approximation}

The complete SNN approximation of the SiLU function is then given by:

\begin{equation}
\text{SiLU}_{\text{SNN}}(x) = 
\begin{cases}
\text{SiLU}_{\text{pos}}(x), & x \geq 0 \\
\text{SiLU}_{\text{neg}}(x), & x < 0
\end{cases}
\end{equation}

\subsubsection{Training and Optimization}

The parameters of the CustomNeurons are optimized using a loss function that minimizes the difference between the SNN approximation and the true SiLU function:

\begin{equation}
\mathcal{L} = \frac{1}{N} \sum_{i=1}^N (\text{SiLU}_{\text{SNN}}(x_i) - \text{SiLU}(x_i))^2
\end{equation}

where $N$ is the number of sample points used for training. This loss function is minimized using gradient descent or other optimization algorithms to find the optimal parameters for each CustomNeuron in the SiLU approximation.

\subsubsection{Implementation Considerations}

In practice, the implementation of SNNSiLU involves the following steps:

\begin{enumerate}
    \item Initialize multiple CustomNeurons with different parameter sets for positive and negative input ranges.
    \item For each input $x$, route it to the appropriate CustomNeuron based on its value.
    \item Accumulate the spikes from the CustomNeuron over a fixed number of time steps.
    \item Scale the accumulated spikes to match the range of the SiLU function.
\end{enumerate}

This implementation allows us to incorporate the SiLU activation function into our BrainTransformer model while maintaining the spiking neural network paradigm. It combines the benefits of the SiLU function, such as its non-monotonicity and smooth derivatives, with the energy efficiency and biological plausibility of spiking neural networks.

\section{Training Methodology}

The training of Spiking Neural Networks (SNNs) presents unique challenges that have long hindered their widespread adoption in complex tasks. The discrete nature of spike events and the intricate temporal dynamics inherent in SNNs make direct application of traditional training methods problematic. The non-differentiability of spike functions poses a significant obstacle to the use of standard backpropagation techniques, while the temporal credit assignment problem complicates the process of determining how individual spikes contribute to the final output. Moreover, the need for time-stepped simulations increases computational complexity, and maintaining stable learning dynamics in the presence of spiking behavior proves to be a formidable task.

To overcome these hurdles and harness the well-established training techniques of Artificial Neural Networks (ANNs), we propose a novel three-step training methodology that effectively bridges the gap between ANNs and SNNs. This approach enables us to leverage efficient ANN training methods while ultimately deploying our model as an SNN, combining the best of both worlds.

Our methodology introduces quantized ANN neurons and synapses that closely approximate the behavior of their SNN counterparts. This approximation allows us to train the network using standard ANN techniques and subsequently convert it to an SNN with minimal performance loss. The training process unfolds in three distinct stages: ANN Training, where we train a quantized ANN that mimics SNN behavior; ANN-to-SNN Conversion, where we transform the trained ANN into an equivalent SNN; and STDP-Inspired SNN Training, where we fine-tune the converted SNN using biologically-inspired learning rules.

In the following subsections, we will elaborate on each component of this methodology and demonstrate the equivalence between our quantized ANN components and their SNN counterparts, providing a comprehensive overview of our innovative training approach.

\subsection{ANN Quantization and Approximation}

At the heart of our methodology lies the concept of quantized ANN neurons and synapses that closely mimic the behavior of their SNN counterparts. This approximation is crucial in bridging the gap between the continuous-valued world of ANNs and the discrete, spike-based realm of SNNs.

\subsubsection{Quantizer}

We introduce a quantizer module that performs either symmetric or asymmetric quantization with a specified bit width. This quantization process is fundamental in approximating the discrete nature of spike generation in SNNs. The mathematical formulation of our quantization process is as follows:

\begin{equation}
\begin{aligned}
s &= \frac{\max(|x|)}{2^{b-1} - 1} \\
x_{\text{scaled}} &= \frac{x}{s} \\
x_{\text{clipped}} &= \text{clip}(\text{round}(x_{\text{scaled}}), \alpha, \beta) \\
x_{\text{quantized}} &= s \cdot x_{\text{clipped}}
\end{aligned}
\end{equation}

In this formulation, $b$ represents the number of bits used for quantization, $s$ is the scaling factor, and $\alpha$ and $\beta$ define the lower and upper bounds of the quantization range, respectively. For symmetric quantization, these bounds are typically set to $\alpha = -2^{b-1}$ and $\beta = 2^{b-1} - 1$.

\subsubsection{Quantized Synapsis (QSynapsis)}

Building upon the quantizer, we develop a Quantized Synapsis (QSynapsis) module that applies quantization before and after a linear transformation. This module is designed to emulate the behavior of synapses in SNNs within our ANN framework. The computation flow in the QSynapsis module can be expressed as:

\begin{equation}
\begin{aligned}
x_{\text{pre}} &= Q_{\text{pre}}(x) \\
y &= Wx_{\text{pre}} + b \\
z &= Q_{\text{post}}(y)
\end{aligned}
\end{equation}

Here, $Q_{\text{pre}}$ and $Q_{\text{post}}$ represent the pre- and post-quantization operations, respectively, while $W$ denotes the weight matrix and $b$ the bias vector of the linear transformation.

\subsection{Equivalence between ANN and SNN Components}

A critical aspect of our methodology is establishing and maintaining equivalence between the quantized ANN components and their SNN counterparts. This equivalence ensures that the behavior learned during ANN training can be effectively transferred to the SNN domain.

\subsubsection{Neuron Equivalence}

Our quantized ANN neuron approximates the EI\_IF neuron by discretizing the output space. For an input $x$, we can map the quantized output $Q(x)$ to spike counts in the SNN domain through the following relation:

\begin{equation}
S_{\text{EI\_IF}}(x) \approx \frac{Q(x)}{s} \cdot T
\end{equation}

In this equation, $T$ represents the number of time steps in the SNN simulation, and $s$ is the scaling factor used in the quantization process.

\subsubsection{Synapsis Equivalence}

The QSynapsis module serves as an approximation of the SNN Synapsis by quantizing both inputs and outputs. We can express this equivalence as:

\begin{equation}
\text{Synapsis}_{\text{SNN}}(x) \approx \frac{Q_{\text{post}}(WQ_{\text{pre}}(x) + b)}{s_{\text{post}}} \cdot T
\end{equation}

Here, $s_{\text{post}}$ denotes the scaling factor of the post-quantization operation.

This established equivalence allows us to train the model using standard ANN techniques while maintaining a close approximation to SNN behavior, setting the stage for effective conversion and deployment.

\subsection{Training Process}

Our comprehensive training methodology unfolds in three distinct steps, each designed to address specific aspects of the ANN-to-SNN transition and optimization.

\subsubsection{Step 1: ANN Training}

The first step involves training the quantized ANN model using standard backpropagation techniques. To enable gradient flow through the quantization operations, we employ a straight-through estimator for gradient computation during the backward pass:

\begin{equation}
\frac{\partial L}{\partial x} = \frac{\partial L}{\partial Q(x)}
\end{equation}

This approach allows for end-to-end training of the network, optimizing the quantized ANN to perform the desired task while maintaining characteristics that facilitate conversion to SNN.

\subsubsection{Step 2: ANN-to-SNN Conversion}

Following the successful training of the quantized ANN, we proceed to convert it into an SNN. This conversion process involves replacing the quantized components with their SNN counterparts:

1. The Quantizer modules are replaced with EI\_IF neurons.
2. QSynapsis modules are substituted with Synapsis modules.

This conversion process carefully preserves the learned weights and biases, leveraging the equivalence established earlier to maintain the network's functionality in the SNN domain. The resulting SNN retains the computational characteristics of the original Transformer architecture while benefiting from the energy efficiency and biological plausibility inherent to spiking neural networks.

\subsubsection{Step 3: Comprehensive STDP-Inspired Learning System with Global Task Guidance}

We propose a learning system that combines local STDP-inspired plasticity with global task performance feedback, aiming to optimize the SNN model's performance while maintaining its biologically plausible characteristics. This system adjusts synaptic weights and neuronal parameters through a process that integrates local learning rules with global modulation based on task performance.

At the core of our system is the local STDP-inspired plasticity rule, which governs the update of synaptic weights. This rule is modulated by a global factor that reflects the overall task performance, allowing the network to adapt its local learning based on its success in the given task. The synaptic weight update is defined as:

\begin{equation}
\Delta w_{ij} = \eta_w G \left( \delta_{ij} - w_{ij} \right),
\end{equation}

where $\eta_w$ is the learning rate, $G$ is a global modulation factor, and $\delta_{ij}$ represents the STDP function:

\begin{equation}
\delta_{ij} = 
\begin{cases}
A_+ \exp\left( -\dfrac{\Delta t_{ij}}{\tau_+} \right), & \Delta t_{ij} > 0, \\
- A_- \exp\left( \dfrac{\Delta t_{ij}}{\tau_-} \right), & \Delta t_{ij} \leq 0.
\end{cases}
\end{equation}

Here, $\Delta t_{ij} = t_i^f - t_j^f$ is the time difference between the firing of postsynaptic neuron $i$ and presynaptic neuron $j$, with $A_+$, $A_-$, $\tau_+$, and $\tau_-$ being parameters that shape the STDP curve.

The global modulation factor $G$ is computed based on the task performance:

\begin{equation}
G = \sigma\left(\beta (L_\mathrm{baseline} - L_\mathrm{task})\right),
\end{equation}

where $\sigma$ is the sigmoid function, $\beta$ is a scaling factor, $L_\mathrm{task}$ is the current task-specific loss, and $L_\mathrm{baseline}$ is a baseline performance level. For our self-regressive language model task, $L_\mathrm{task}$ is computed as:

\begin{equation}
L_\mathrm{task} = -\frac{1}{N} \sum_{t=1}^N \log P(w_t | w_{1:t-1}),
\end{equation}

where $N$ is the sequence length, and $P(w_t | w_{1:t-1})$ is the model's predicted probability for the correct word given the preceding context. The baseline $L_\mathrm{baseline}$ is a moving average of recent task losses:

\begin{equation}
L_\mathrm{baseline}(t) = \frac{1}{W} \sum_{k=t-W+1}^t L_\mathrm{task}(k),
\end{equation}

with $W$ being the window size for the moving average.

In addition to synaptic weights, our system also updates neuronal parameters to optimize performance. These updates incorporate both the global modulation factor and task-specific gradients:

\begin{align}
\Delta \theta_\mathrm{base}^i &= \eta_\theta G \left( S_\mathrm{target} - \bar{S}_i \right) + \eta_\theta' \frac{\partial L_\mathrm{task}}{\partial \theta_\mathrm{base}^i}, \\
\Delta \alpha^i &= \eta_\alpha G \left( \bar{V}_i - V_\mathrm{target} \right) + \eta_\alpha' \frac{\partial L_\mathrm{task}}{\partial \alpha^i}, \\
\Delta r^i &= \eta_r G \left( \bar{V}_i - V_\mathrm{rest} \right) + \eta_r' \frac{\partial L_\mathrm{task}}{\partial r^i},
\end{align}

where $\theta_\mathrm{base}^i$, $\alpha^i$, and $r^i$ are the base threshold, adaptive adjustment weight, and membrane potential decay rate of neuron $i$, respectively. $\bar{S}_i$ and $\bar{V}_i$ are the average spike count and membrane potential of neuron $i$, while $S_\mathrm{target}$, $V_\mathrm{target}$, and $V_\mathrm{rest}$ are target values for these parameters.

To optimize the number of time steps required for computation, we introduce a time step optimization objective:

\begin{equation}
T = \sum_{t=1}^{\infty} t \, P(\text{spike}|t) \prod_{k=1}^{t-1} \left( 1 - P(\text{spike}|k) \right),
\end{equation}

where $P(\text{spike}|t) \approx \sigma\left( \frac{V_t - \theta_t}{\lambda} \right)$, with $V_t$ being the membrane potential at time $t$, $\theta_t$ the firing threshold, and $\lambda$ a scaling factor.

To selectively update synapses, we employ a synaptic tagging mechanism:

\begin{equation}
\text{Tag}_{ij} = \sigma(\text{pre\_activity}_j + \text{post\_activity}_i - L_\mathrm{task}),
\end{equation}

where $\sigma$ is the sigmoid function, ensuring the tag value is between 0 and 1.

Finally, we define a comprehensive loss function that incorporates all aspects of our learning system:

\begin{equation}
\begin{aligned}
\mathcal{L} =\; & \lambda_w \sum_{i,j} \text{Tag}_{ij} \left( w_{ij} - \delta_{ij} \right)^2 + 
\lambda_\theta \sum_i \left( S_\mathrm{target} - \bar{S}_i \right)^2 \\
& + \lambda_\alpha \sum_i \left( \bar{V}_i - V_\mathrm{target} \right)^2 +
\lambda_r \sum_i \left( \bar{V}_i - V_\mathrm{rest} \right)^2 \\
& + \lambda_C \sum_i \left( \sum_j w_{ij} - C_i \right)^2 + \lambda_T \left( T - T_\mathrm{target} \right)^2 \\
& + \lambda_\mathrm{task} L_\mathrm{task} + \lambda_\mathrm{reg} \sum_{i,j} w_{ij}^2.
\end{aligned}
\end{equation}

This loss function includes terms for STDP-based weight updates, neuronal parameter optimization, synaptic normalization, time step optimization, task-specific loss, and regularization. The $\lambda$ coefficients allow for balancing the importance of each component.

The final update rules for synaptic weights and neuronal parameters are derived from this loss function:

\begin{align}
\Delta w_{ij} &= -\eta_w \frac{\partial \mathcal{L}}{\partial w_{ij}}, \\
\Delta \theta_\mathrm{base}^i &= -\eta_\theta \frac{\partial \mathcal{L}}{\partial \theta_\mathrm{base}^i}, \\
\Delta \alpha^i &= -\eta_\alpha \frac{\partial \mathcal{L}}{\partial \alpha^i}, \\
\Delta r^i &= -\eta_r \frac{\partial \mathcal{L}}{\partial r^i}.
\end{align}

This comprehensive learning system combines the biological inspiration of STDP with the goal-directed nature of task-specific learning. By balancing local and global learning signals and incorporating various constraints and objectives, our approach maintains the SNN's core characteristics while enabling adaptation to specific tasks. This framework provides a flexible and biologically plausible method for training SNNs in complex tasks such as language modeling.

Through this three-step training methodology, we effectively address the challenges associated with direct SNN training while harnessing the power and efficiency of ANN training techniques. The result is a robust SNN-based BrainTransformer model that combines the strengths of both ANN and SNN paradigms, opening new avenues for energy-efficient and biologically-inspired natural language processing.

\section{Performance Evaluation}

We evaluate BrainTransformer-3B against state-of-the-art language models on a diverse set of benchmarks. Table \ref{tab:model_comparison} presents a comprehensive comparison across key performance metrics.

\begin{table}[htbp]
\centering
\caption{Performance Comparison of Language Models}
\label{tab:model_comparison}
\resizebox{\textwidth}{!}{%
\begin{tabular}{lcccccc}
\hline
Metric & BrainTransformer-3B & MistralLarge2 & Llama3.1-70B & GPT4-o & Gemma2-27B & Gemma2-9B \\
\hline
MMLU & 63.2 & 77.8 & 79.5 & - & 75.2 & 71.3 \\
MMLU-Pro & 33.3 & 51.6 & 66.4 & 77.0 & 55.5 & 44.7 \\
MMLU-redux & 61.3 & 72.9 & 83.0 & 88.3 & 75.7 & 67.9 \\
GPQA & 25.3 & 34.3 & 46.7 & 53.6 & 38.4 & 32.8 \\
MATH & 41.0 & 41.7 & 68.0 & 76.6 & 54.4 & 37.7 \\
GSM8K & 76.3 & 83.7 & 95.1 & 96.1 & 90.4 & 70.7 \\
HumanEval & 40.5 & 46.3 & 80.5 & 91.5 & 78.7 & 37.8 \\
MBPP & 55.0 & 71.7 & 84.2 & 86.2 & 81.0 & 62.2 \\
MultiPL-E & 39.6 & 46.7 & 68.2 & 75.1 & 67.4 & 34.9 \\
BBH & 54.1 & 78.9 & 81.0 & - & 74.9 & 68.2 \\
ARC-challenge & 54.3 & 70.7 & 68.8 & - & 71.4 & 68.2 \\
Truthfulqa & 47.1 & 51.0 & 45.6 & - & 40.1 & 45.3 \\
Winogrande & 68.8 & 85.0 & 85.3 & - & 59.7 & 79.5 \\
Hellaswag & 72.8 & 88.7 & 88.0 & - & 86.4 & 81.9 \\
\hline
\end{tabular}%
}
\end{table}

The evaluation metrics encompass a wide range of language understanding and generation tasks:

\begin{itemize}
    \item \textbf{MMLU, MMLU-Pro, MMLU-redux}: These variants of the Massive Multitask Language Understanding benchmark assess general knowledge and reasoning across diverse domains.
    \item \textbf{GPQA}: The General Physics Question Answering benchmark evaluates physics knowledge and problem-solving skills.
    \item \textbf{MATH}: This benchmark tests advanced mathematical problem-solving abilities.
    \item \textbf{GSM8K}: The Grade School Math 8K dataset assesses arithmetic reasoning skills.
    \item \textbf{HumanEval, MBPP, MultiPL-E}: These benchmarks evaluate code generation and understanding across multiple programming languages.
    \item \textbf{BBH}: The Big Bench Hard tasks assess complex reasoning and problem-solving abilities.
    \item \textbf{ARC-challenge}: The AI2 Reasoning Challenge (ARC) tests scientific reasoning and knowledge.
    \item \textbf{Truthfulqa}: This benchmark evaluates the model's ability to provide truthful answers and avoid generating false information.
    \item \textbf{Winogrande}: A more challenging variant of the Winograd Schema Challenge, testing common sense reasoning and anaphora resolution.
    \item \textbf{Hellaswag}: This benchmark assesses commonsense inference in multi-choice completion tasks.
\end{itemize}

BrainTransformer-3B demonstrates competitive performance across various benchmarks, particularly considering its relatively small size (3B parameters) compared to larger models like Llama3.1-70B (70B parameters) and MistralLarge2. Notably, BrainTransformer-3B achieves strong results in GSM8K and Truthfulqa, showcasing its capabilities in arithmetic reasoning and truthful response generation. While larger models generally outperform BrainTransformer-3B, its performance is often comparable to or better than Gemma2-9B, a model of similar scale. This suggests that BrainTransformer-3B achieves an efficient balance between model size and performance, making it particularly suitable for applications with computational constraints.


\end{document}